\documentclass[]{gtech}

\usepackage{amssymb}
\usepackage{multirow}
\usepackage{bigdelim}
\usepackage{todonotes}
\usepackage{longtable}
\usepackage{tabularray}
\usepackage{wrapfig}
\usepackage[most]{tcolorbox} 
\usepackage{xcolor}
\usepackage{url}
\usepackage{algorithm}  
\usepackage{algpseudocode} 
\usepackage[toc,page]{appendix}
\usepackage{multirow}
\usepackage[normalem]{ulem}
\usepackage{adjustbox}
\usepackage{booktabs}  
\usepackage{natbib}
\usepackage{colortbl}  
\usepackage{subcaption}
\usepackage{pifont}
\usepackage{amsmath}
\usepackage{amsfonts}       
\usepackage{multirow}
\usepackage{mathrsfs}
\usepackage{tabularx}
\usepackage[utf8]{inputenc} 
\usepackage[T1]{fontenc}    
\usepackage{hyperref}       
\usepackage{url}            
\usepackage{booktabs}       
\usepackage{nicefrac}       
\usepackage{microtype}      
\usepackage{xcolor}         
\definecolor{ourreward}{HTML}{32bbee}
\definecolor{sparsereward}{HTML}{ed9a81}
\usepackage{graphicx}
\usepackage{wrapfig}
\usepackage{cleveref}
\usepackage{fvextra}
\tcbuselibrary{listings, breakable}

\usepackage{bbding}
\usepackage{perpage}
\MakePerPage{footnote}
\usepackage{caption}
\usepackage{CJK}

\usepackage{bbm}
\useunder{\uline}{\ul}{}

\title{DR-Venus: Towards Frontier Edge-Scale Deep Research Agents with Only 10K Open Data}

\author{Venus Team, Ant Group}

\abstract{
\vspace{-1em}
\begin{center}
\bfseries Abstract
\end{center}
\vspace{-0.5em}
Edge-scale deep research agents based on small language models are attractive for real-world deployment due to their advantages in cost, latency, and privacy. In this work, we study how to train a strong small deep research agent under limited open-data by improving both data quality and data utilization. We present \textbf{DR-Venus}, a frontier 4B deep research agent for edge-scale deployment, built entirely on open data.
Our training recipe consists of two stages. In the first stage, we use agentic supervised fine-tuning (SFT) to establish basic agentic capability, combining strict data cleaning with resampling of long-horizon trajectories to improve data quality and utilization. In the second stage, we apply agentic reinforcement learning (RL) to further improve execution reliability on long-horizon deep research tasks. To make RL effective for small agents in this setting, we build on IGPO and design turn-level rewards based on information gain and format-aware regularization, thereby enhancing supervision density and turn-level credit assignment.
Built entirely on roughly 10K open-data, DR-Venus-4B significantly outperforms prior agentic models under 9B parameters on multiple deep research benchmarks, while also narrowing the gap to much larger 30B-class systems. Our further analysis shows that 4B agents already possess surprisingly strong performance potential, highlighting both the deployment promise of small models and the value of test-time scaling in this setting.  We release our models, code, and key recipes to support reproducible research on edge-scale deep research agents.
}

\gtechdata[Code]{\url{https://github.com/inclusionAI/DR-Venus}}
\gtechdata[Model]
{\url{https://huggingface.co/collections/inclusionAI/dr-venus}}

\begin{document}
\maketitle



\begin{figure}[h]
    \centering
    \vspace{1em}
    \includegraphics[width=\linewidth]{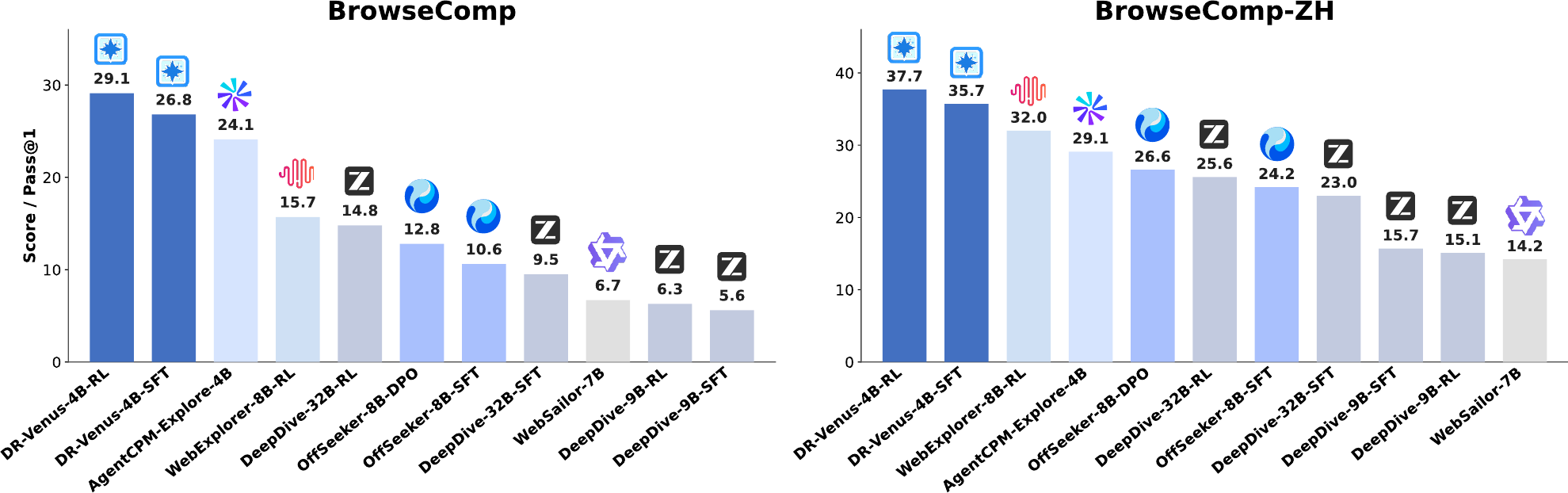}
    \caption{
    Performance comparison of DR-Venus-4B against other open-source models on BrowseComp and BrowseComp-ZH.
    }
    \label{fig:hero_intro}
\end{figure}

\newpage
\section{Introduction}

Recent advances in large language models have enabled increasingly capable agents that can reason, use tools, and interact with external environments~\citep{qin2024tool, qu2025tool}. Among these emerging systems, deep research agents have attracted growing attention for their ability to solve complex information-seeking tasks through iterative search, browsing, evidence collection, and answer synthesis over long horizons~\citep{shi2025deep}. Because such tasks require sustained planning, reliable tool use, and long-horizon evidence integration, deep research provides a particularly challenging and representative testbed for agentic language models~\citep{wei2025browsecomp}.

At the same time, bringing deep research capability to edge-scale small language models is highly desirable for real-world deployment~\citep{chen2026agentcpm, yang2026nanbeige4}. Small models offer practical advantages in cost, latency, and privacy, making them an attractive foundation for lightweight research assistants that can be deployed more broadly. However, most existing deep research systems are built on substantially larger models and often rely on closed data or relatively complex training pipelines~\citep{team2025mirothinker, team2025tongyi, tang2025beyond, chu2026redsearcher, du2026openseeker}. As a result, the frontier of small deep research agents under an open-data setting remains insufficiently explored.

In this work, we study a central question in this setting: \emph{how can we train a strong small deep research agent even under limited open-data supervision?} We argue that this is fundamentally a problem of improving both \emph{data quality} and \emph{data utilization}. On the one hand, small models are more sensitive to noisy trajectories, formatting artifacts, and imperfect tool-use traces, making data quality especially important. On the other hand, the limited capability of small models makes agentic RL particularly challenging: rollout groups often contain no successful trajectories at all, leading to advantage collapse and severely reduced training efficiency. This makes effective supervision allocation crucial, especially toward behaviors most critical for long-horizon agentic search.

Motivated by this perspective, we propose \textbf{DR-Venus}, a 4B frontier edge-scale deep research agent trained entirely on open data. Our training recipe consists of two stages. In the first stage, we perform agentic supervised fine-tuning (SFT) on cleaned and resampled REDSearcher~\citep{chu2026redsearcher} trajectories to establish basic deep research capability. In the second stage, we apply agentic reinforcement learning (RL) to further improve execution reliability in long-horizon tasks. To make RL effective for small agents, we build on IGPO~\citep{wang2026information} and develop turn-level rewards tailored to long-horizon deep research, combining information-gain supervision with format penalties to improve supervision density, credit assignment, and behavioral stability during training.

Built entirely on roughly 10K open-data, DR-Venus-4B-SFT already establishes a remarkably strong 4B baseline and consistently outperforms prior agentic systems under 9B parameters on most benchmarks. With agentic RL, DR-Venus-4B-RL further improves over the SFT baseline and sets a new frontier among small agents. Despite its small size, DR-Venus also narrows the gap to substantially larger 30B-class systems, suggesting that careful improvement of data quality and effective data utilization can compensate for a substantial portion of the scale gap.

Moreover, our analyses provide several additional insights. First, long-horizon trajectory resampling substantially improves the effectiveness of limited open-data SFT supervision. Second, turn-level RL with IGPO is much more effective than conventional sparse trajectory-level optimization for long-horizon deep research. Third, Pass@K evaluation reveals that the capability ceiling of small deep research agents is already surprisingly high, suggesting that test-time scaling may be especially effective for unlocking the potential of small reasoning models. Finally, tool-use analysis shows that successful trajectories consistently rely more on browsing than failed ones, and that RL further calibrates tool use toward more effective evidence acquisition.

To facilitate transparent and reproducible progress in this direction, we release our models, code, and key recipes. We hope this work provides a practical starting point for future research on edge-scale deep research agents built from small models and open data.
\section{Methodology}
In this section, we present the overall training recipe of DR-Venus. We first formulate the deep research task as a long-horizon reasoning-and-acting problem over external environments, where the agent must iteratively reason, invoke tools, gather evidence, and produce a final answer (Section~\ref{sec:formulation}). Under limited open-data supervision, our training recipe is designed around two complementary goals: improving training data quality and data utilization.
To this end, we adopt a two-stage pipeline. In the first stage, we use agentic SFT on cleaned and resampled trajectories to build basic agentic capability (Section~\ref{sec:sft}). In the second stage, we further apply agentic RL with turn-level credit assignment to improve execution reliability and push performance toward the frontier (Section~\ref{sec:rl}).

\subsection{Problem Formulation}~\label{sec:formulation}

We formulate a \emph{Deep Research Agent} as a language model-based policy that solves complex information-seeking tasks through long-horizon interaction with an external environment. Given a user query $q$, the agent must iteratively reason, search for relevant information, inspect retrieved content, collect intermediate evidence, and finally synthesize an answer. Unlike standard question answering, the agent cannot rely solely on parametric knowledge; instead, it must actively acquire and integrate external information over multiple interaction turns.

Formally, we consider an environment $\mathcal{E}$ equipped with a set of executable actions $\mathcal{A}$, including \textbf{\texttt{search}}, \textbf{\texttt{browse}}, and \textbf{\texttt{answer}} actions. Since the agent in our setting is built upon a \emph{reasoning model}, each interaction turn consists of not only an action, but also an intermediate reasoning process. We initialize the interaction history with the user query, i.e., $h_0=q$. At turn $t$, the agent generates a turn output $u_t=(\tau_t,a_t)$ conditioned on the interaction history $h_{<t}$, where $\tau_t$ denotes the model's intermediate reasoning at turn $t$, and $a_t \in \mathcal{A}$ is the corresponding tool or answer action:
\[
u_t=(\tau_t,a_t)\sim\pi_{\theta}(\cdot|h_{\le t-1}).
\]
When $a_t$ is a tool action, the environment returns an observation $o_t$, and the history is updated to
\[
h_{\le t}=(h_{\le t-1},u_t,o_t).
\]
When $a_t$ is the answer action, the interaction ends at turn $T$, and no further environment observation is returned. The resulting interaction trajectory is
\[
H=(q, (u_1,o_1), (u_2,o_2),...,(u_{T-1},o_{T-1}),u_T).
\]

Under this formulation, solving a deep research task requires the agent to generate a trajectory that both gathers sufficient external evidence and supports a correct final answer.

Following this formulation, our goal is to learn a policy that can both (1) acquire basic agentic capability from supervised trajectories, and (2) improve long-horizon execution reliability through reinforcement learning. In the first agentic SFT stage, the model learns from high-quality open-data trajectories containing reasoning, actions, and observations. In the second agentic RL stage, the model is further optimized using turn-level and trajectory-level rewards that encourage correct answers, stable tool use, and consistent progress over long interaction horizons.

\subsection{Building Basic Agentic Capability with Supervised Fine-Tuning} \label{sec:sft}

To equip the model with basic agentic capability, we perform supervised fine-tuning (SFT) on REDSearcher~\citep{chu2026redsearcher} trajectories. Because the raw trajectories contain substantial redundancy, structural mismatch, and noisy supervision, we first apply a multi-stage data filtering and construction pipeline to obtain higher-quality training data. We then fine-tune the model on the resulting trajectories, masking out all non-assistant tokens in the training loss.

\subsubsection{Data Filtering and Construction}~\label{sec:data_clean}
We build the SFT data from 10{,}001 raw REDSearcher trajectories~\footnote{\url{https://huggingface.co/datasets/Zchu/REDSearcher_SFT_10K}}. The data construction pipeline includes four steps:

\begin{itemize}
\item \textbf{Environment alignment.}
We first convert all trajectories into the same interaction format used by our online inference pipeline, including the message schema, system prompt, tool-call arguments, and tool-response format. This reduces the mismatch between training and inference and ensures that the model learns tool use under the same protocol as deployment. All 10,001 raw trajectories are successfully converted in this stage.

\item \textbf{Disallowed-tool pruning and duplicate removal.}
Our runtime environment only exposes two tools, \texttt{search} and \texttt{browse}. Therefore, we remove all other tool interactions at the turn level, together with their paired tool-response turns, rather than discarding the entire trajectory. This affects 1,064 trajectories and removes 3,378 disallowed tool calls, mostly from \texttt{PythonInterpreter}. We then remove duplicate \texttt{search} and \texttt{browse} calls and their paired responses. Duplicate tool calls are common, appearing in 6,821 trajectories, and we remove 15,728 such duplicated interactions in total. Most of them are duplicate \texttt{browse} events, indicating that redundancy mainly arises during webpage inspection rather than search. After these cleaning steps, 10,000 trajectories remain valid.

\item \textbf{Correctness filtering.}
We further apply correctness filtering after structural cleaning, retaining only trajectories whose final answers are judged to be correct. We use \texttt{Qwen3-235B-A22B-Instruct-2507} as the judge model and adapt the judge implementation from Tongyi-DeepResearch~\citep{team2025tongyi}. This step retains 9,365 trajectories, corresponding to 93.65\% of the valid set.

\item \textbf{Turn-aware resampling.} 
Since long-horizon deep research requires sustained planning, repeated tool interaction, and multi-step evidence aggregation, we upweight longer trajectories during SFT through turn-aware resampling. Specifically, we assign sampling weights of 1$\times$, 2$\times$, and 5$\times$ to trajectories with 0--50, 51--100, and more than 100 turns, respectively. This increases the final training set from 9,365 to 18,745 instances, while raising the proportion of trajectories longer than 50 turns from 60.28\% to 80.15\% and that of trajectories longer than 100 turns from 13.29\% to 33.21\%.
\end{itemize}

Overall, this pipeline improves both SFT data quality and data utilization: it removes structurally invalid or low-value supervision, and it increases the utilization of limited open-data trajectories by emphasizing long-horizon interactions that are most relevant to deep research.

\subsubsection{Agentic Supervised Fine-Tuning}

Following the trajectory formulation in Section~\ref{sec:formulation}, we use agentic SFT to initialize the model with basic deep research capability before reinforcement learning. In our setting, the role of agentic SFT is not only to teach the model the interaction pattern of reasoning, tool use, and answer generation, but also to make effective use of limited open-data supervision. This is particularly important for edge-scale small models, which are more sensitive to noisy trajectories, inconsistent tool formatting, and imperfect long-horizon behaviors.

Concretely, each trajectory is serialized into a single autoregressive sequence, and we optimize the standard next-token prediction objective on the assistant-generated tokens, including reasoning traces, tool invocations, and the final answer, while masking environment observation tokens from the loss. Formally, given the cleaned SFT dataset $\mathcal{D}_{\mathrm{SFT}}$, the training objective is
\begin{equation}
\mathcal{L}_{\mathrm{SFT}}(\theta)
=
-\sum_{H \in \mathcal{D}_{\mathrm{SFT}}}
\sum_{i \in \mathcal{M}(H)}
\log \pi_\theta(x_i \mid x_{<i}),
\label{eq:sft_loss}
\end{equation}
where $\mathcal{M}(H)$ denotes the agent-generated token positions in the serialized trajectory $H$, corresponding to the reasoning traces $\tau_t$ and actions $a_t$ in each turn $t$, while excluding observations $o_t$ returned from external environments.

In summary, agentic SFT provides the model with a stable initialization for structured tool use and long-horizon interaction, while concentrating supervision on the most reliable and informative parts of the limited open-data corpus.

\subsection{Pushing Toward Frontier Performance with Reinforcement Learning} \label{sec:rl}

Even after agentic SFT, DR-Venus still exhibits several failure modes that compromise execution reliability, including formatting errors, redundant reasoning, and inefficient tool use. To address these issues, we further train DR-Venus with agentic reinforcement learning (RL). However, high-quality open-source training data for Agentic RL is extremely scarce and expensive to construct. To address this challenge, our solution is to compensate for the lack of data volume by building denser reward signals that improve data efficiency. Specifically, we adopt Information Gain-based Policy Optimization (IGPO)~\citep{wang2026information}, a simple yet effective agentic RL algorithm that leverages the idea of information gain to construct dense turn-level reward signals. Notably, IGPO adopts a GRPO~\citep{shao2024deepseekmath}-style rollout scheme, in which a group of rollouts $\{H_i\}_{i=1}^G$ is sampled for each query. This design can also improve data efficiency to some extent.
\subsubsection{Turn-Level Reward Design for Long-Horizon Agentic Tasks}
\textbf{Information Gain Reward.}
To enable dense turn-level rewards for improved data efficiency, IGPO~\citep{wang2026information} formulates deep research as a process of iteratively collecting information relevant to the ground truth. Under this view, each policy-generated turn can be evaluated by how much new evidence it contributes toward identifying the correct answer. This formulation naturally defines the value of each turn as the extent to which it increases the policy’s probability of generating the ground truth, i.e., the information gain with respect to the ground truth. Following IGPO~\citep{wang2026information}, we refer to this turn-level value signal as the information gain (IG) reward, since it characterizes how much the current turn improves the model’s confidence in the correct answer.

We denote the ground truth token sequence by $g=(g_1,\ldots,g_L)$. At turn $t$ in rollout $H_i$, the log probability assigned to $g$ by the current policy $\pi_\theta$ is given by
\begin{equation}
\label{eq:logp}
\log \pi_\theta(g \mid h_{i,\leq t})
=
\frac{1}{L}\sum_{j=1}^{L}
\log \pi_\theta(g_j\mid h_{i,\leq t},g_{<j})
,
\end{equation}
where $h_{i,\leq t}$ denotes the interaction history in rollout $H_i$ up to and including turn $t$. Then the IG reward~\footnote{Because the IG reward is derived from log probabilities, we apply a stop-gradient operation to it during optimization.} for turn $t$ is defined as
\begin{equation}
\label{eq:info gain reward}
r_{i,t}^{IG}
=\log \pi_\theta(g \mid h_{i,\leq t})-\log \pi_\theta(g \mid h_{i,\leq t-1}),\qquad 1\leq t<T.
\end{equation} 
For implementation consistency, the ground-truth answer $g$ is cast into the same structured format as a sampled model response in the rollout, e.g., \texttt{<think>Now there's enough information to answer</think><answer>Ground Truth $g$</answer>}. Notably, the IG reward is not applied to the answer turn. Instead, the answer turn retains the outcome reward based on answer correctness, such as rule-based rewards or an LLM-judge reward, denoted by $r_i^{\text{O}}$.

\textbf{Browse-Aware IG Assignment.}
In deep research tasks, browse turns typically expose the policy to more concrete evidence, whereas search turns are primarily exploratory and may introduce noisy or weakly informative intermediate states. Motivated by this distinction, we optionally adopt a browse-aware IG assignment strategy: we compute IG rewards only on browse turns, and assign each such reward to the browse turn itself and all preceding search turns since the previous browse.

\textbf{Turn-Level Format Penalty.} As noted earlier, agentic SFT equips the model with basic agentic capabilities, but residual instability in format correctness still harms execution reliability. This makes format-aware reward design necessary. Nevertheless, traditional trajectory-level format penalties apply the same penalty across all tokens in a trajectory, which may unfairly penalize correctly formatted turns due to errors elsewhere. For long-horizon deep research tasks, which may involve more than 200 turns, such coarse-grained regulation is clearly inadequate. We therefore introduce a turn-level format penalty that enables more fine-grained control over formatting behavior, which is combined with the IG reward and the outcome reward as follows.

\begin{equation}
\hat{r}_{i,t}=
\begin{cases}
r_{i,t}, & \text{if the format at turn } t \text{ is valid},\\
-\lambda_{\mathrm{fmt}}, & \text{otherwise},
\end{cases}
\end{equation}
where $r_{i,t}$ denotes either $r_{i,t}^{IG}$ or $r_i^{\text{O}}$, and $-\lambda_{\mathrm{fmt}}$ is the format penalty coefficient. The original reward is preserved for correctly formatted turns, and replaced by $-\lambda_{\mathrm{fmt}}$ otherwise. This allows the penalty to target malformed turns precisely while avoiding noisy supervision.

\textbf{Normalization and Discounted Accumulation.}
For each rollout group containing $G$ trajectories $\{H_i\}_{i=1}^G$, trajectory $H_i$ consists of $T_i$ turns and yields a sequence of format-adjusted IG rewards $\{\hat{r}_{i,t}^{\mathrm{IG}}\}_{t=1}^{T_i-1}$ together with a format-adjusted outcome reward $\hat{r}_{i}^{\mathrm{O}}$. Since turn-level IG rewards and outcome rewards are defined at different granularities and may have different scales, we normalize them separately within each rollout group to improve reward balancing and stabilize optimization:
\begin{equation}
\label{eq:norm}
\tilde{r}_{i,t} =
\begin{cases}
\dfrac{\hat{r}_{i,t}^{\mathrm{IG}} - \mu^{\mathrm{IG}}}{\sigma^{\mathrm{IG}}}, & 1 \le t < T_i, \\[8pt]
\dfrac{\hat{r}_{i}^{\mathrm{O}} - \mu^{\mathrm{O}}}{\sigma^{\mathrm{O}}}, & t = T_i,
\end{cases}
\end{equation}
where $\mu^{\mathrm{IG}}$ and $\sigma^{\mathrm{IG}}$ are the mean and standard deviation of the format-adjusted IG rewards $\{\hat{r}_{i,t}^{\mathrm{IG}}\}$ within the rollout group, and $\mu^{\mathrm{O}}$ and $\sigma^{\mathrm{O}}$ are defined analogously for the format-adjusted outcome rewards $\{\hat{r}_{i}^{\mathrm{O}}\}$. For convenience, we denote the normalized terminal outcome reward by $\tilde{r}_{i}^{\mathrm{O}} := \tilde{r}_{i,T_i}$.

For ultra-long-horizon deep research tasks (e.g., 200 turns of tool-use), outcome supervision can be extremely weak, as correct final answers are often rare and the outcome rewards within a rollout group may collapse to, or be dominated by, zeros. In this regime, optimization may be overly driven by IG rewards, increasing the risk of converging to a local optimum. To address this issue, we introduce an optional rescaling strategy, termed \textbf{IG-Scale}, which adaptively adjusts the magnitude of normalized IG rewards based on the scale of the outcome rewards, thereby improving reward balance between turn-level IG supervision and terminal outcome supervision.

Specifically, we compute the batch-level average absolute magnitudes of normalized outcome rewards and normalized IG rewards, where $B$ denotes the batch size:
\begin{equation}
M^{\mathrm{O}}=\frac{1}{B}\sum_{i=1}^{B}\left|\tilde{r}_{i}^{\mathrm{O}}\right|,\qquad
M^{\mathrm{IG}}=\frac{1}{\sum_{i=1}^{B}(T_i-1)}\sum_{i=1}^{B}\sum_{t=1}^{T_i-1}\left|\tilde{r}_{i,t}^{\text{IG}}\right|.
\end{equation}
The scaling factor is then defined as
\begin{equation}
s=\min\!\left(\frac{\max(M^{\mathrm{O}},\eta)}{M^{\mathrm{IG}}+\delta},\, s_{\max}\right),
\qquad \eta=0.3,\; \delta=10^{-8},\; s_{\max}=10.
\end{equation}
We define the final turn-level reward used for optimization as
\begin{equation}
\label{eq:ig_scale}
\bar{r}_{i,t}=
\begin{cases}
\operatorname{IG\mbox{-}Scale}(\tilde{r}_{i,t}; s)=s\cdot \tilde{r}_{i,t}, & 1 \le t < T_i, \\[8pt]
\tilde{r}_{i,t}, & t = T_i.
\end{cases}
\end{equation}
If IG-Scale is disabled, we set $\bar{r}_{i,t}=\tilde{r}_{i,t}$. Intuitively, IG-Scale rebalances IG and outcome signals by suppressing IG influence when outcome supervision is weak. It is optional and trades off update aggressiveness for robustness: without it, updates are more aggressive; with it, updates are more conservative by aligning the IG scale with the outcome scale.

Regardless of whether IG-Scale is applied, the resulting turn-level reward captures only the immediate consequence of the current decision, rather than its effect on future turns. To account for such long-term dependencies, we introduce a turn-level discounted cumulative reward $\tilde{R}_{i,t}$:
\begin{equation}
\label{eq:disc-adv}
\tilde{R}_{i,t}=\sum_{k=t}^{T_i}\gamma^{\,k-t}\bar{r}_{i,k},
\end{equation}
where $\gamma\in[0,1]$ denotes the discount factor. During optimization, $\tilde{R}_{i,t}$ is assigned to every token in $u_{i,t}$, the policy-generated output at turn $t$ of rollout $H_i$, thereby providing a dense supervision signal that incorporates future reward information.

\subsubsection{Policy Optimization via IGPO}

Let $u_i=(u_{i,1},\dots,u_{i,|u_i|})$ denote the policy-generated tokens in rollout $H_i$, and let $c_{i,k}$ denote the full serialized prefix of $u_{i,k}$, including environment-returned observations, as the token-level form of the interaction history defined in Section~\ref{sec:formulation}. For tokens from turn $t$, we set $\tilde{R}_{i,k}=\tilde{R}_{i,t}$.

Building on this token-level reward assignment, we retain the GRPO-style objective with turn-level credit assignment. Formally, the IGPO objective~\citep{wang2026information} is
\begin{equation}
\label{eq:igpo}
\begin{aligned}
\mathcal{J}_{\mathrm{IGPO}}(\theta) 
=\; &
\mathbb{E}_{\{H_i\}_{i=1}^{G}}
\Bigg[
\frac{1}{G}\sum_{i=1}^{G}\frac{1}{|u_i|}
\sum_{k=1}^{|u_i|}
\min\!\Bigg(
\frac{\pi_\theta(u_{i,k}\mid c_{i,k})}
{\pi_{\theta_{\mathrm{old}}}(u_{i,k}\mid c_{i,k})}
\,\tilde{R}_{i,k}, \\[4pt]
& \qquad\qquad
\mathrm{clip}\!\left(
\frac{\pi_\theta(u_{i,k}\mid c_{i,k})}
{\pi_{\theta_{\mathrm{old}}}(u_{i,k}\mid c_{i,k})},
\,1-\epsilon,\,1+\epsilon
\right)\tilde{R}_{i,k}
\Bigg)
-\beta\,\mathbb{D}_{\mathrm{KL}}(\pi_\theta \,\|\, \pi_{\mathrm{ref}})
\Bigg],
\end{aligned}
\end{equation}
where $\epsilon$ is the clipping threshold, and $\beta$ controls the strength of the KL penalty.
\section{Experiments}

\subsection{Experimental Settings}

\textbf{Training Data.} We use two types of training data. For agentic SFT, we construct the trajectory corpus based on the open-source REDSearcher trajectories~\footnote{\url{https://huggingface.co/datasets/Zchu/REDSearcher_SFT_10K}}, after applying the cleaning and turn-aware resampling procedure described in Section~\ref{sec:data_clean}. For agentic RL, we use 1k query-answer pairs curated from the REDSearcher data source~\footnote{\url{https://huggingface.co/datasets/Zchu/REDSearcher_RL_1K}}. Thus, both stages of DR-Venus are trained entirely on open data, with agentic SFT focusing on trajectory imitation and agentic RL focusing on optimization over curated QA supervision.

\textbf{Benchmarks.} We evaluate DR-Venus on six typical benchmarks covering deep research, web browsing, and multi-step information-seeking: (1) \textbf{BrowseComp}~\citep{wei2025browsecomp}, which evaluates long-horizon web browsing and information-seeking in English; (2) \textbf{BrowseComp-ZH}~\citep{zhou2025browsecomp}, which extends BrowseComp setting to the Chinese web; (3) \textbf{GAIA (Text-Only)}~\citep{mialon2023gaia}, a benchmark for general AI assistants with tasks that often require multi-step search and reasoning, where we use the text-only subset; (4) \textbf{xBenchDS-2505} and \textbf{xBenchDS-2510}~\citep{chen2025xbench}, two versions of xbench-DeepSearch used to evaluate deep research ability under the xbench framework; and (5) \textbf{DeepSearchQA}~\citep{gupta2026deepsearchqa}, a benchmark designed to evaluate multi-step deep research with an emphasis on comprehensive answer generation. For datasets containing fewer than 300 examples, we report the mean performance over three independent evaluation runs to reduce variance. For the remaining datasets, we report the results from a single evaluation run.

\textbf{Baselines.} We compare DR-Venus with three groups of baselines.
(1) \textbf{Frontier foundation models with tools}, including GLM-4.7~\citep{zeng2025glm}, MiniMax-M2.1, DeepSeek-V3.2~\citep{liu2025deepseek}, Kimi-K2.5~\citep{team2026kimi}, Claude-4.5-Opus, OpenAI-o3, GPT-5 High, and Gemini-3-Pro.
(2) \textbf{Open-source trained agents at larger scales ($\geq$ 30B)}, including 
DeepDive-32B~\citep{lu2025deepdive},
SMTL-30B-300~\citep{chen2026search}, WebSailor-V2-30B~\citep{li2025websailorv2}, Tongyi-DR-30B~\citep{team2025tongyi}, DeepMiner-32B-RL~\citep{tang2025beyond}, OpenSeeker-v1-30B-SFT~\citep{du2026openseeker}, OpenResearcher-30B-A3B~\citep{li2026openresearcher} and REDSearcher-30B-A3B~\citep{chu2026redsearcher}.
(3) \textbf{Open-source small agents ($\leq$ 9B)}, including DeepDive-9B, WebSailor-7B~\citep{li2025websailor}, OffSeeker-8B~\citep{zhou2026offseeker}, WebExplorer-8B-RL~\citep{liu2025webexplorer} and AgentCPM-Explore-4B~\citep{chen2026agentcpm}.
For fair comparison, we do not compare with methods that rely on additional context management or test-time scaling techniques beyond the standard evaluation setting, such as RE-TRAC-4B~\citep{zhu2026re}, Marco-DR-8B~\citep{zhu2026marco}, and MiroThinker-v1.0~\citep{team2025mirothinker}.

\textbf{Tool Server.} Following standard practice~\citep{team2025tongyi,chu2026redsearcher}, we equip the agent with two external tools: \texttt{search} and \texttt{browse}. The \texttt{search} tool is powered by \texttt{Serper}~\footnote{\url{https://serper.dev/}}, which interfaces with the Google Search API and returns the top-10 search results for each query. The \texttt{browse} tool is built on top of \texttt{Jina}~\footnote{\url{https://jina.ai/}} for web page reading. To support content summarization within the \texttt{browse} tool, we further deploy \texttt{Qwen3-30B-A3B-Instruct-2507} as the summarization model.

\textbf{Implementation Details.}
We use \texttt{Qwen3-4B-Thinking-2507}~\footnote{\url{https://huggingface.co/Qwen/Qwen3-4B-Thinking-2507}} as the backbone. Agentic SFT is conducted on 8 A100 GPUs with the verl~\footnote{\url{https://github.com/verl-project/verl}} FSDP trainer. The maximum training length is set to 200K tokens, with a global batch size of 32, a per-GPU micro-batch size of 1, and a learning rate of 1e-5. We train for one epoch with multi-turn supervision, right truncation, gradient checkpointing, and sequence parallelism of size 8. Subsequently, Agentic RL is executed on 16 A100 GPUs utilizing the verl vLLM engine and the FSDP trainer. For this phase, the training batch size is set to 16. During the rollout phase, the maximum context length is up to 256K tokens, with a maximum generation length of 8,192 tokens per turn. The group size is set to 8, and the temperature is configured at 1.0. For IGPO-specific settings, we enable \textit{browse-aware IG assignment} and \textit{IG-Scale}, set $\lambda_{\mathrm{fmt}}=1.0$, and use a discount factor of $\gamma=0.95$. During evaluation, we allow at most 200 interaction steps per query. Decoding uses temperature 1.0, top-$p$ 0.95, top-$k$ 20, and a presence penalty of 1.1, with a maximum token budget of 256K. For all benchmarks, we follow the official evaluation protocols and adopt the provided evaluation prompts whenever available. The full system prompt of our inference pipeline is included in Appendix~\ref{sec:prompt}.

\subsection{Main Results}

\begin{table*}[t]
\centering
\caption{Overall performance comparison on six widely used deep research benchmarks.}
\label{tab:main-results}
\setlength{\tabcolsep}{4pt}
\renewcommand{\arraystretch}{1.0}
\resizebox{0.9\textwidth}{!}{
\begin{tabular}{l c c c c c c}
\toprule
\textbf{Model} &
\begin{tabular}{@{}c@{}}\textbf{Browse}\\\textbf{Comp}\end{tabular} &
\begin{tabular}{@{}c@{}}\textbf{Browse}\\\textbf{Comp-ZH}\end{tabular} &
\begin{tabular}{@{}c@{}}\textbf{GAIA}\\\textbf{(Text-Only)}\end{tabular} &
\begin{tabular}{@{}c@{}}\textbf{xBench-}\\\textbf{DS-2505}\end{tabular} &
\begin{tabular}{@{}c@{}}\textbf{xBench-}\\\textbf{DS-2510}\end{tabular} &
\begin{tabular}{@{}c@{}}\textbf{Deep}\\\textbf{SearchQA}\end{tabular} \\
\midrule

\multicolumn{7}{l}{\textbf{\shortstack[l]{Foundation Models}}} \\
\midrule
GLM-4.7 & 67.5 & 66.6 & 61.9 & 72.0 & 52.3 & -- \\
MiniMax-M2.1 & 62.0 & 47.8 & 64.3 & 68.7 & 43.0 & -- \\
DeepSeek-V3.2 & 67.6 & 65.0 & 75.1 & 78.0 & 55.7 & 60.9 \\
Kimi-K2.5 & 74.9 & 62.3 & -- & -- & 46.0 & 77.1 \\
Claude-4.5-Opus & 67.8 & 62.4 & -- & -- & -- & 80.0 \\
OpenAI-o3 & 49.7 & 58.1 & -- & 67.0 & -- & -- \\
GPT-5 High & 54.9 & 65.0 & 76.4 & 77.8 & 75.0 & 79.0 \\
Gemini-3-Pro & 59.2 & 66.8 & -- & -- & 53.0 & 76.9 \\
\midrule
\multicolumn{7}{l}{\textbf{\shortstack[l]{Trained Agents ($\geq$30B)}}} \\
\midrule
DeepDive-32B-SFT & 9.5 & 23.0 & -- & 48.5 & -- & -- \\
DeepDive-32B-RL & 14.8 & 25.6 & -- & 50.5 & -- & -- \\
SMTL-30B-300 & 48.6 & -- & 75.7 & 82.0 & -- & -- \\
WebSailor-V2-30B-RL & 35.3 & 44.1 & 74.1 & 73.7 & -- & -- \\
Tongyi-DR-30B & 43.4 & 46.7 & 70.9 & 75.0 & 55.0 & -- \\
DeepMiner-32B-RL & 33.5 & 40.1 & 58.7 & 62.0 & -- & -- \\
OpenSeeker-v1-30B-SFT & 29.5 & 48.4 & -- & 74.0 & -- & -- \\
OpenResearcher-30B-A3B & 26.3 & -- & 64.1 & 65.0 & -- & -- \\
REDSearcher-30B-A3B  & 42.1 & 49.8 & 80.1 & -- & -- & -- \\
\midrule
\multicolumn{7}{l}{\textbf{\shortstack[l]{Trained Agents ($\leq$9B)}}} \\
\midrule
DeepDive-9B-SFT & 5.6 & 15.7 & -- & 35.0 & -- & -- \\
DeepDive-9B-RL & 6.3 & 15.1 & -- & 38.0 & -- & -- \\
WebSailor-7B & 6.7 & 14.2 & 37.9 & 34.3 & -- & -- \\
OffSeeker-8B-SFT & 10.6 & 24.2 & 47.6 & 48.0 & -- & -- \\
OffSeeker-8B-DPO & 12.8 & 26.6 & 51.5 & 49.0 & -- & -- \\
WebExplorer-8B-RL & 15.7 & 32.0 & 50.0 & 53.7 & 23.0 & 17.8 \\
AgentCPM-Explore-4B & 24.1 & 29.1 & 63.9 & 70.0 & 34.0 & 32.8  \\
\rowcolor{blue!10} \textbf{DR-Venus-4B-SFT} & 26.8 & 35.7 & 65.4 & 69.0 & 35.3 & 37.7 \\
\rowcolor{blue!10} \textbf{DR-Venus-4B-RL} & 29.1 & 37.7 & 64.4 & 74.7 & 40.7 & 39.6 \\
\bottomrule
\end{tabular}
}
\end{table*}

Table~\ref{tab:main-results} reports the overall performance of DR-Venus and representative baselines on six widely used deep research benchmarks. Several observations can be drawn from these results:

First, even with supervised fine-tuning alone, \textbf{DR-Venus-4B-SFT} already establishes a remarkably strong small-model baseline. It consistently outperforms previously reported 4B--9B agentic systems on most benchmarks. In particular, compared with AgentCPM-Explore-4B, DR-Venus-4B-SFT achieves clear gains on BrowseComp (\textbf{+2.7}), BrowseComp-ZH (\textbf{+6.6}), GAIA (\textbf{+1.5}), xBench-DS-2510 (\textbf{+1.3}), and DeepSearchQA (\textbf{+4.9}). These results show that a carefully designed open-data training recipe is already sufficient to push a 4B model to a substantially stronger frontier.

More importantly, agentic RL further unlocks the potential of the small model. With agentic RL, \textbf{DR-Venus-4B-RL} establishes a new state of the art among small deep research agents and improves over the SFT baseline on five of the six benchmarks, with gains of \textbf{+2.3} on BrowseComp, \textbf{+2.0} on BrowseComp-ZH, \textbf{+5.7} on xBench-DS-2505, \textbf{+5.4} on xBench-DS-2510, and \textbf{+1.9} on DeepSearchQA. These results show that once basic agentic capability is initialized by SFT, agentic RL is crucial for converting that capability into stronger and more reliable long-horizon performance. Our analysis finds that these improvements are closely associated with better formatting accuracy, more stable tool use, and stronger execution reliability over long interaction trajectories.

The comparison with larger trained agents is also encouraging. Despite being substantially smaller than 30B-scale systems, DR-Venus-4B-SFT already matches or exceeds several of them on individual benchmarks, and DR-Venus-4B-RL further narrows the gap to much larger agents. For example, DR-Venus-4B-SFT already surpasses OpenResearcher-30B-A3B on all reported benchmarks, while DR-Venus-4B-RL further strengthens this advantage. On xBench-DS-2505, DR-Venus-4B-RL reaches \textbf{74.7}, approaching Tongyi-DR-30B (\textbf{75.0}). These results suggest that, under limited open-data supervision, careful improvement of data quality and effective data utilization can compensate for a substantial portion of the scale gap. In other words, strong deep research performance is not determined by model scale alone. With a well-designed open-data pipeline and effective RL, even a 4B model can reach a highly competitive frontier.

\subsection{Ablation Study}


\begin{table}[t]
\centering
\caption{Ablation study on BrowseComp and BrowseComp-ZH. All SFT models are trained based on REDSearcher trajectories.}
\label{tab:sft_init_results}
\begin{tabular}{lccc}
\toprule
\textbf{Model} & \textbf{Training} & \textbf{BrowseComp} & \textbf{BrowseComp-ZH} \\
\midrule
REDSearcher-30B-A3B (SFT) & SFT & 34.7 & 26.8 \\
\hline
DR-Venus-4B-SFT (w/o Resampling) & SFT & 22.8 & 33.9 \\
DR-Venus-4B-SFT (w/ Resampling, Ours) & SFT & 26.8 (\textcolor{red}{+4.0}) & 35.7 (\textcolor{red}{+1.8}) \\
\hline
DR-Venus-4B-RL (w/ GRPO) & SFT+RL & 25.3 (\textcolor{blue}{-1.5}) & 35.6 (\textcolor{blue}{-0.1}) \\
DR-Venus-4B-RL (w/ IGPO, Ours) & SFT+RL & 29.1 (\textcolor{red}{+2.3}) & 37.7 (\textcolor{red}{+2.0}) \\
\bottomrule
\end{tabular}
\end{table}

Table~\ref{tab:sft_init_results} reports an ablation study of our training pipeline on BrowseComp and BrowseComp-ZH. The results show that even without agentic RL, DR-Venus-4B-SFT already acquires strong deep research capability from the open REDSearcher trajectories. This suggests that high-quality agentic SFT is sufficient to provide a competitive small-model initialization for long-horizon reasoning, tool use, and evidence aggregation.

The comparison between DR-Venus-4B-SFT (w/o Resampling) and DR-Venus-4B-SFT (w/ Resampling) further highlights the importance of effective data utilization under limited open-data supervision. Both models are trained on the same underlying trajectory source, and the only difference is whether long-horizon trajectories are upweighted during SFT. The resampled version generally performs better, improving BrowseComp by \textbf{+4.0} and BrowseComp-ZH by \textbf{+1.8}. This shows that redistributing supervision toward long-horizon interaction traces is an effective way to further improve small-model deep research capability.

This result is especially notable given the comparison with \textbf{REDSearcher-30B-A3B}. Despite being much smaller, \textbf{DR-Venus-4B-SFT} already surpasses REDSearcher-30B-A3B on BrowseComp-ZH. This suggests that, under limited open-data supervision, model scale alone is not the determining factor. Instead, how the data is cleaned, aligned, and utilized can matter just as much, if not more. In other words, the open REDSearcher trajectories still contain \textit{substantial untapped value}, and a carefully designed SFT pipeline can unlock much stronger performance even with a 4B model.

Agentic RL further improves this strong SFT baseline and pushes DR-Venus toward a new small-model frontier. Compared with DR-Venus-4B-SFT (w/ Resampling), the two RL variants exhibit clearly different behaviors: GRPO brings little to no improvement, whereas IGPO consistently yields additional gains on both benchmarks. Specifically, on BrowseComp, DR-Venus-4B-RL (w/ IGPO) improves over the SFT baseline by \textbf{+2.3}, while GRPO leads to a decrease of \textbf{-1.5}; on BrowseComp-ZH, the gain is \textbf{+2.0} for IGPO, whereas GRPO is nearly flat at \textbf{-0.1}. These results suggest that obtaining consistent gains from agentic RL is itself nontrivial for long-horizon deep research tasks (e.g., trajectories with around 200 turns). Instead, the effectiveness of RL depends critically on whether the reward design provides sufficiently dense and well-aligned turn-level supervision. In our setting, IGPO is substantially more effective than conventional sparse trajectory-level optimization, making it a more suitable RL strategy for stabilizing and improving small-model agent behavior.

\subsection{Analysis of Capability Boundary}

\begin{figure}[t]
    \centering
    \includegraphics[width=0.48\linewidth]{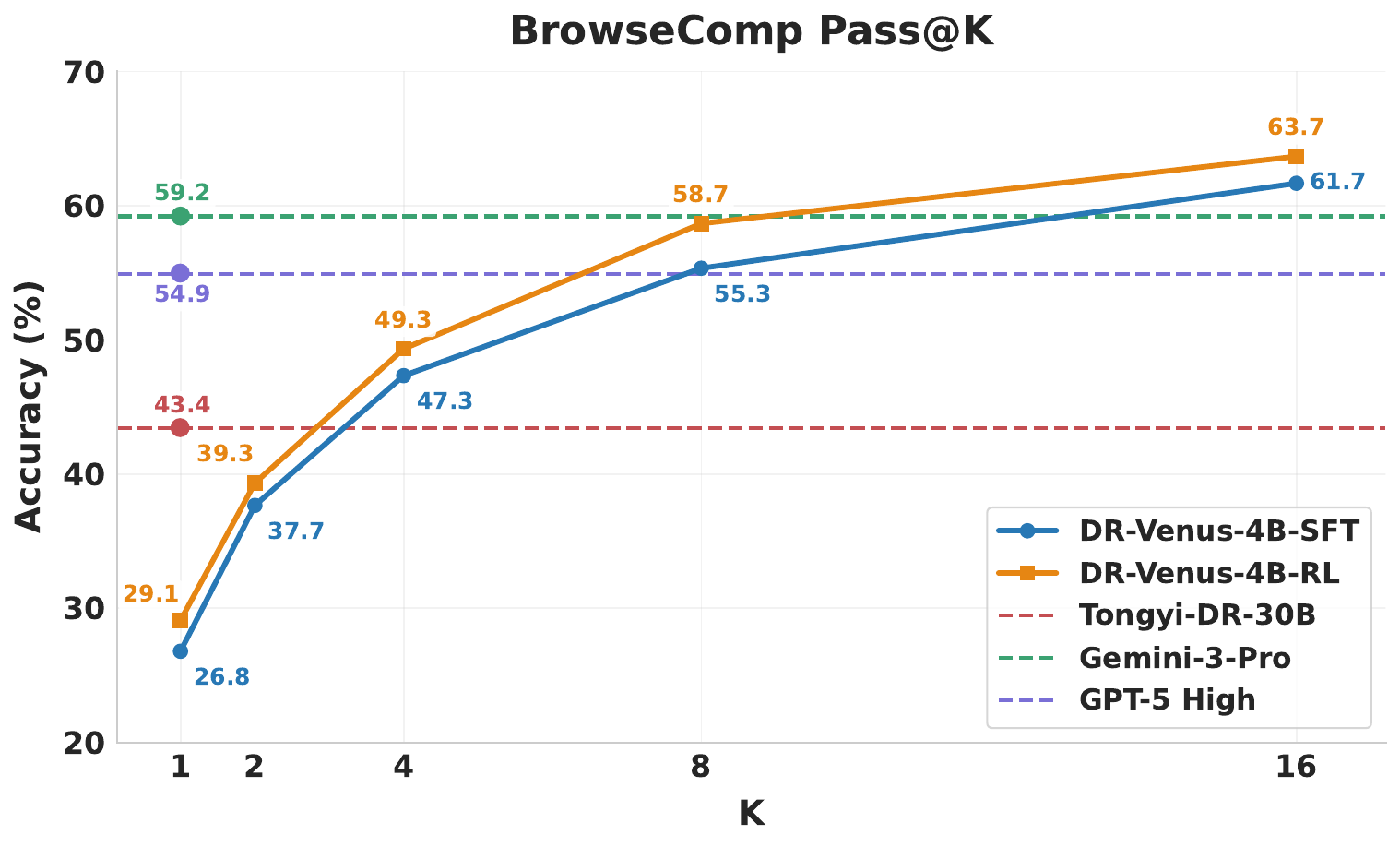}
    \hfill
    \includegraphics[width=0.48\linewidth]{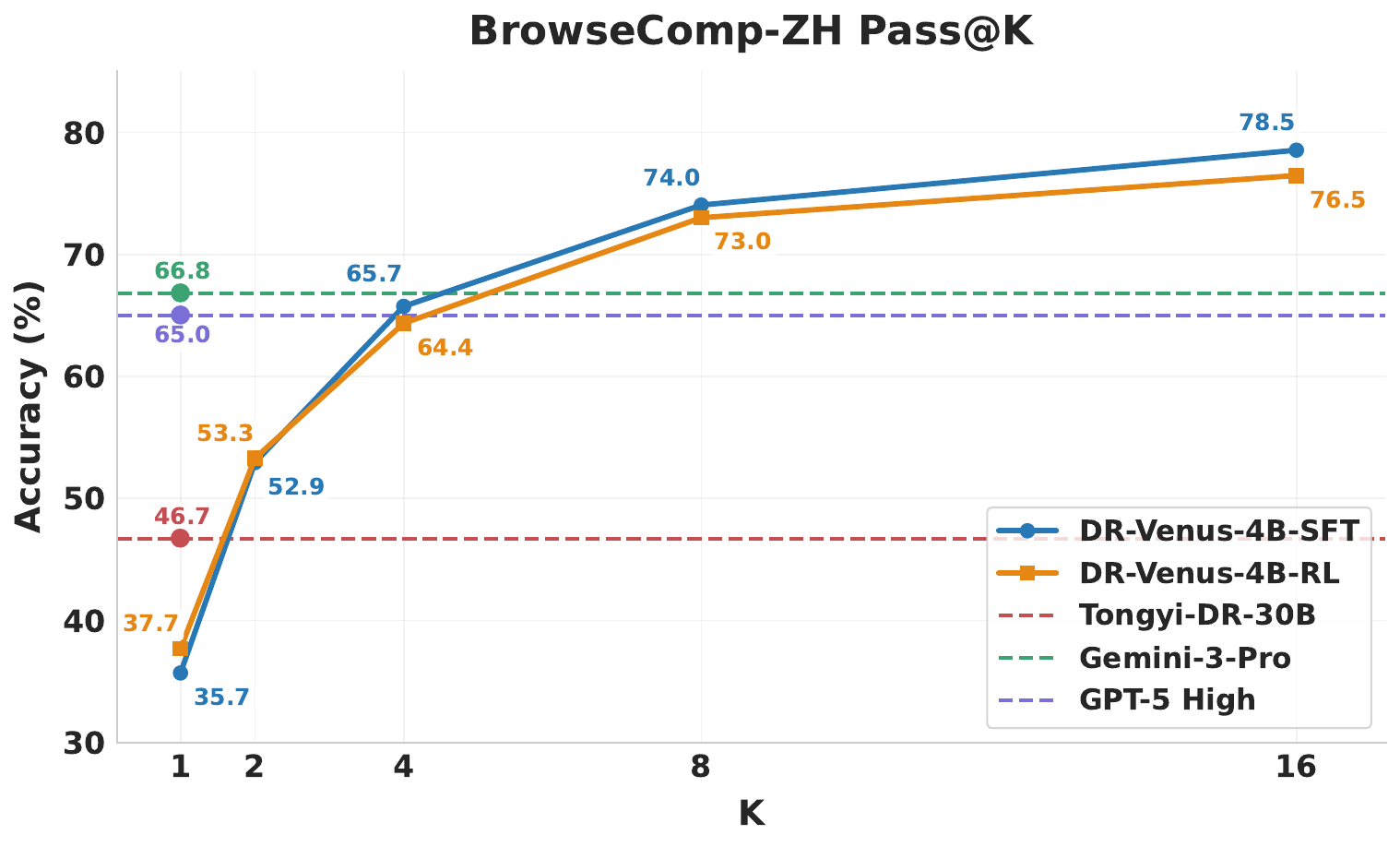}
    \caption{Pass@K performance of DR-Venus on BrowseComp (left) and BrowseComp-ZH (right). }
    \label{fig:passk_rl}
\end{figure}

We further analyze the capability boundary of DR-Venus by comparing SFT and RL checkpoints using Pass@K on BrowseComp and BrowseComp-ZH. Figure~\ref{fig:passk_rl} reveals a clear and consistent pattern: the capability ceiling of small deep research agents is already surprisingly high, while RL mainly improves the reliability of realizing this capability under limited sampling budgets.

On BrowseComp, DR-Venus-4B-RL outperforms DR-Venus-4B-SFT across all values of $K$, improving from 26.8 to 29.1 at Pass@1 and from 61.7 to 63.7 at Pass@16. This suggests that RL not only improves low-budget success, but also pushes the overall capability boundary upward to some extent. In contrast, on BrowseComp-ZH, the largest gains from RL appear at small values of $K$: Pass@1 improves from 35.7 to 37.7, and Pass@2 from 52.9 to 53.3. However, at larger values of $K$, the SFT model already achieves very strong results, reaching 74.0 at Pass@8 and 78.5 at Pass@16, while the RL model is slightly lower at 73.0 and 76.5, respectively. This indicates that, at least in the current setting, RL does not necessarily raise the ultimate ceiling on every benchmark; instead, its main effect is often to make strong trajectories emerge more reliably in the low-$K$ regime. The weaker large-$K$ performance on BrowseComp-ZH may partly reflect a distribution mismatch, since our current RL training data is entirely English.

These results suggest that the latent capability of small deep research agents may be substantially underestimated when evaluated only with Pass@1. Once \textbf{test-time scaling} is allowed, a 4B model can already reach a very strong performance frontier. This is especially evident on BrowseComp-ZH, where DR-Venus-4B-SFT reaches 78.5 at Pass@16, significantly exceeding Tongyi-DR-30B (46.7) and even outperforming strong proprietary foundation models such as Gemini-3-Pro (66.8) and GPT-5 High (65.0). These findings highlight the large deployment potential of edge-scale deep research agents and suggest that test-time scaling may be an especially effective way to unlock the capability of small reasoning models.

\begin{figure}[t]
    \centering
    \vspace{1em}
    \includegraphics[width=\linewidth]{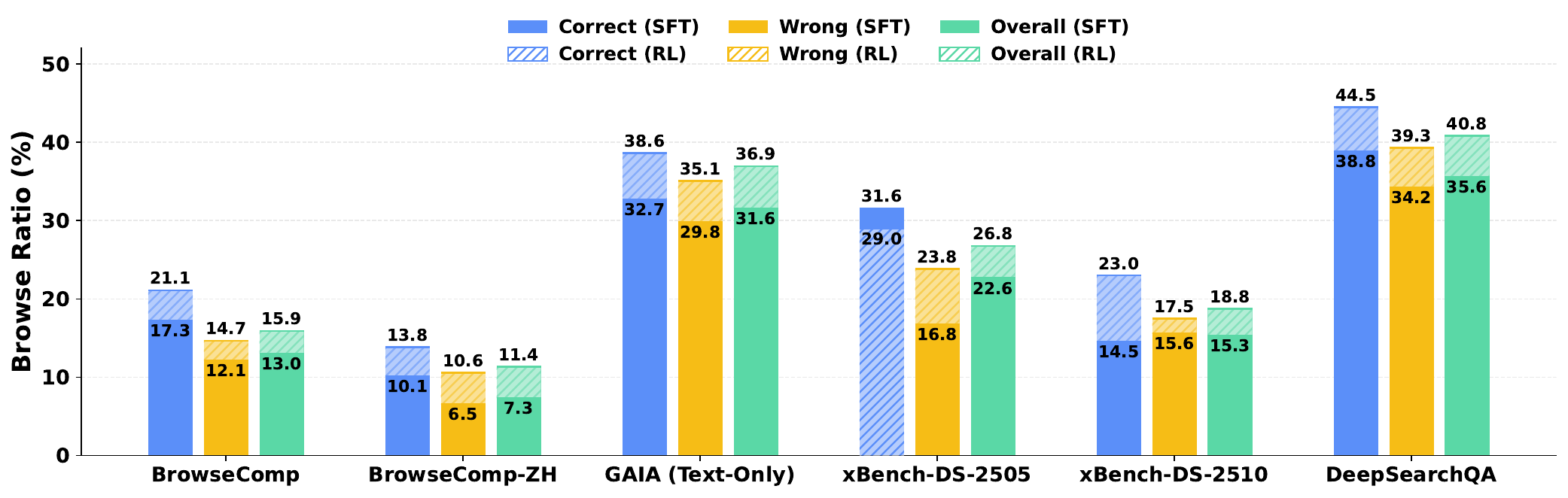}
    \caption{Browse ratio of correct, wrong, and overall trajectories under SFT and RL across six benchmarks.}
    \label{fig:tool_use_ratio}
\end{figure}

\subsection{Analysis of Tool Use}

Figure~\ref{fig:tool_use_ratio} shows the browse ratio of correct, wrong, and overall trajectories under SFT and RL across six benchmarks. A clear pattern emerges across almost all settings: \textbf{correct trajectories consistently exhibit a higher browse ratio than wrong trajectories}. This trend already appears in the SFT model and becomes even more pronounced after RL. The result suggests that, for complex deep research tasks, search alone is often insufficient. \texttt{Search} mainly provides short snippets, whereas \texttt{browse} allows the agent to inspect the underlying webpages and gather the detailed evidence needed for grounded reasoning. Therefore, successful trajectories are more likely to move beyond shallow retrieval and perform deeper evidence inspection.

RL further strengthens this behavior by increasing the overall browse ratio and better aligning tool use with task success. At the aggregate level, the overall browse ratio rises from \textbf{17.49\%} under SFT to \textbf{22.46\%} under RL, while the browse ratio of correct trajectories increases from \textbf{23.71\%} to \textbf{28.96\%}. More importantly, RL makes the ``correct $>$ wrong'' pattern more consistent across benchmarks. For example, on xBench-DS-2510, SFT shows a slightly counterintuitive pattern where wrong trajectories browse more than correct ones (15.57\% vs. 14.51\%), whereas RL reverses this relation (22.99\% vs. 17.50\%). This suggests that RL does not simply suppress ineffective search or encourage more browsing overall, but instead steers tool use toward better evidence gathering.

\section{Conclusion}
In this work, we present DR-Venus, a frontier 4B edge-scale deep research agent built entirely on open data. Our method follows a simple two-stage recipe: agentic SFT with strict trajectory cleaning and long-horizon resampling, followed by agentic RL with IGPO algorithm. DR-Venus-4B significantly outperforms prior agentic models at a similar scale and narrows the gap to substantially larger 30B-class systems. Our results further show that, under limited open-data supervision, careful improvement of training data quality and effective data utilization is sufficient to unlock strong deep research capability in small models, while RL is essential for stabilizing formatting, tool use, and long-horizon execution. We hope our released models, code, and recipes can provide a practical starting point for future research on edge-scale deep research agents.


\section{Contributions}
All contributors of DR-Venus are listed in alphabetical order by their last names.

\subsection{Core Contributors}
Sunhao Dai$^{\dag}$, Yong Deng, Jinzhen Lin, Yusheng Song, Guoqing Wang, Xiaofeng Wu, Yuqi Zhou

\subsection{Contributors}
Shuo Yang, Zhenzhe Ying, Zhanwei Zhang

\subsection{Supervisors}
Changhua~Meng$^{\dag}$, Weiqiang~Wang

{\let\thefootnote\relax\footnote{$^{\dag}$Corresponding Authors: Sunhao~Dai(sunhaodai@gmail.com), Changhua~Meng(changhua.mch@antgroup.com).}}

\newpage
\bibliographystyle{abbrvnat}
\bibliography{main}

\newpage

\begin{appendix}

\section{System Prompt}~\label{sec:prompt}
In this section, we present the complete system prompts used in DR-Venus. Our prompts are adapted from those of REDSearcher~\citep{chu2026redsearcher} and Tongyi DeepResearch~\citep{team2025tongyi}, and are further refined to better support the reasoning and interaction patterns.
\begin{tcolorbox}[
  title=System Prompt of DR-Venus,
  colback=blue!2,
  colframe=blue!40!black,
  coltitle=white,
  colbacktitle=blue!60!black,
  fonttitle=\bfseries,
  boxrule=0.6pt,
  arc=1mm,
  listing only,  
  listing engine=listings,
  breakable
]
You are a deep research assistant. Your core function is to conduct thorough, multi-source investigations into any topic. You must handle both broad, open-domain inquiries and queries within specialized academic fields. For each user request, you must actively seek out and **cross-check information** from credible and diverse sources, then integrate the findings into a response that is comprehensive, accurate, well-structured, and objective. When you have gathered sufficient information and are ready to provide the definitive response, you must enclose the entire final answer in \textbf{<answer></answer>} tags.\\

\# Tools\\

You may call one or more functions to assist with the user query.\\

You are provided with function signatures within <tools></tools> XML tags:

\textbf{<tools>}

\begin{Verbatim}[
  breaklines=true,
  breaksymbolleft={}
]
{"type": "function", "function": {"name": "search", "description": "Perform Google web searches then returns a string of the top search results. Accepts multiple queries.", "parameters": {"type": "object", "properties": {"query": {"type": "array", "items": {"type": "string", "description": "The search query."}, "minItems": 1, "description": "The list of search queries."}}, "required": ["query"]}}}
\end{Verbatim}
\begin{Verbatim}[
  breaklines=true,
  breaksymbolleft={}
]
{"type": "function", "function": {"name": "visit", "description": "Visit webpage(s) and return the summary of the content.", "parameters": {"type": "object", "properties": {"url": {"type": "array", "items": {"type": "string"}, "description": "The URL(s) of the webpage(s) to visit. Can be a single URL or an array of URLs."}, "goal": {"type": "string", "description": "The specific information goal for visiting webpage(s)."}}, "required": ["url", "goal"]}}}
\end{Verbatim}
\textbf{</tools>}\\

For each function call, return a json object with function name and arguments within \texttt{\textbf{<tool\_call></tool\_call>}} XML tags:

\texttt{\textbf{<tool\_call>}}

\begin{verbatim}
{"name": <function-name>, "arguments": <args-json-object>}
\end{verbatim}

\texttt{\textbf{</tool\_call>}}\\

Current date: 2026-03-01
\end{tcolorbox}

\end{appendix}

\end{document}